%% file: SW.tex
\documentclass[10pt,twocolumn,letterpaper]{article}

\usepackage{iccv}
\usepackage{times}
\usepackage{epsfig}
\usepackage{graphicx}
\usepackage{amsmath}
\usepackage{amssymb}
\usepackage{nccmath}

\usepackage{makecell}
\usepackage{xcolor}
\usepackage{multirow}
\usepackage{fixltx2e}
\usepackage{bm}

\usepackage[pagebackref=true,breaklinks=true,letterpaper=true,colorlinks,bookmarks=false]{hyperref}
\usepackage{algpseudocode}
\usepackage{algorithm}

\newcommand{\R}{\rotatebox[origin=c]{90}}

\iccvfinalcopy 

\def\iccvPaperID{382} 

\ificcvfinal\fi
\begin{document}
	
	\title{Switchable Whitening for Deep Representation Learning}	
	\author{
		\setlength{\parskip}{1.5\baselineskip}
		Xingang Pan\textsuperscript{1}, 
		Xiaohang Zhan\textsuperscript{1}, 
		Jianping Shi\textsuperscript{2}, 
		Xiaoou Tang\textsuperscript{1},
		and Ping Luo\textsuperscript{1,3}
		\vspace{0.2cm}
		\\
		\normalsize
		\textsuperscript{1}{CUHK-SenseTime Joint Lab, The Chinese University of Hong Kong} \\
		\normalsize
		\textsuperscript{2}{SenseTime Group Limited} \
		\quad\textsuperscript{3}{The University of Hong Kong} \\
		\tt\small
		\{px117, zx017, xtang, pluo\}@ie.cuhk.edu.hk, \
		shijianping@sensetime.com
	}
	
	\maketitle

	\input{./sections/abstract.tex}

	\input{./sections/introduction.tex}
	\input{./sections/relatedwork.tex}
	\input{./sections/SW.tex}

\input{./sections/experiments.tex}
	\input{./sections/conclusion.tex}
	
	{\small
		\bibliographystyle{ieee}
		\bibliography{egbib}
	}

	\input{./sections/supplementary.tex}
	
\end{document}

%% file: sections/abstract.tex
\begin{abstract}
	
	Normalization methods are essential components in convolutional neural networks (CNNs).
	%
	They either \textit{standardize} or \textit{whiten} data using statistics estimated in predefined sets of pixels.
	Unlike existing works that design normalization techniques for specific tasks, we propose Switchable Whitening (SW), which provides a general form unifying different whitening methods as well as standardization methods. 
	SW learns to switch among these operations in an end-to-end manner.
	%
	It has several advantages.
	First, SW adaptively selects appropriate whitening or standardization statistics for different tasks (see Fig.1), making it well suited for a wide range of tasks without manual design.
	%
	Second, by integrating the benefits of different normalizers, SW shows consistent improvements over its counterparts in various challenging benchmarks.
	Third, SW serves as a useful tool for understanding the characteristics of whitening and standardization techniques.
	
	We show that SW outperforms other alternatives on image classification (CIFAR-10/100, ImageNet), semantic segmentation (ADE20K, Cityscapes), domain adaptation (GTA5, Cityscapes), and image style transfer (COCO).
	For example, without bells and whistles, we achieve state-of-the-art performance with 45.33\% mIoU on the ADE20K dataset.
	Code is available at \url{https://github.com/XingangPan/Switchable-Whitening}.

\end{abstract}

%% file: sections/introduction.tex
\section{Introduction}

\begin{figure}[!t]
	\centering
	\includegraphics[width=8.6cm]{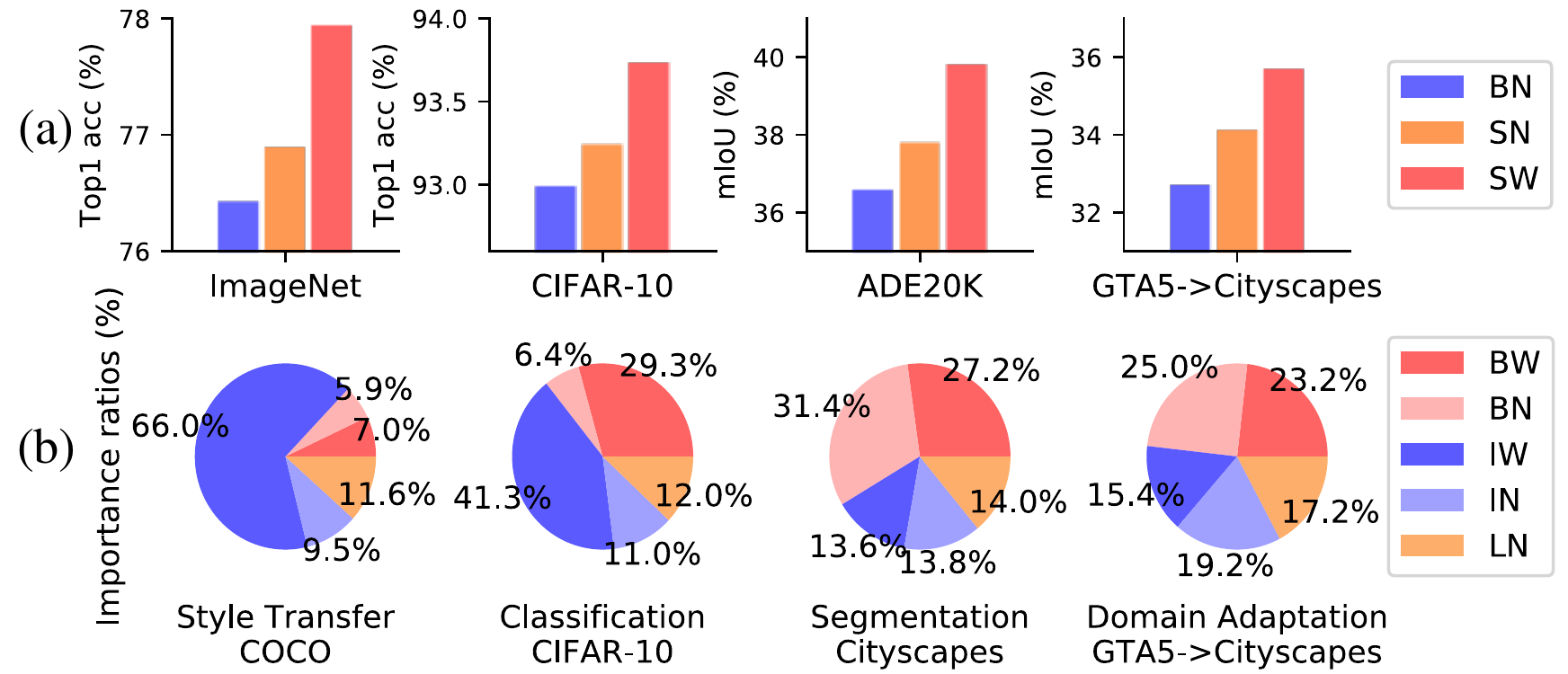}
	\vspace{-0.3cm}
	\caption{\label{Fig1} (a) SW outperforms its counterparts in a variety of benchmarks. (b) SW learns to select appropriate whitening or standardization methods in different tasks and datasets. The CNNs are ResNet50 for ImageNet and ADE20K, ResNet44 for CIFAR-10, and VGG16 for GTA5$\rightarrow$Cityscapes.
		GTA5$\rightarrow$Cityscapes indicates adapting from GTA5 to Cityscapes using domain adaptation.} 
	\vspace{-0.3cm}
\end{figure}
%

Normalization methods have been widely used as a basic module in convolutional neural networks (CNNs).
In various applications, different normalization techniques like Batch Normalization (BN)~\cite{ioffe2015batch}, Instance Normalization (IN)~\cite{ulyanov2017improved} and Layer Normalization (LN)~\cite{ba2016layer} are proposed.
These normalization techniques generally perform \textit{standardization} that \textit{centers} and \textit{scales} features.
Nevertheless, the features are not \textit{decorrelated}, hence their correlation still exists.

Another type of normalization methods is \textit{whitening}, which not only standardizes but also \textit{decorrelates} features. 
For example, Decorrelated Batch Normalization (DBN)~\cite{lei2018decorrelated}, or namely Batch Whitening (BW), whitens a mini-batch using its covariance matrix, which gives rise to better optimization efficiency than BN in image classification.
Moreover, whitening features of an individual image is used in image style transfer~\cite{li2017universal} to filter out information of image appearance.
Here we refer to this operation as instance whitening (IW).
Despite their successes, existing works applied these whitening techniques separately to different tasks, preventing them from benefiting each other.
%
%
Besides, whitening and standardization methods are typically employed in different layers of a CNN, which complicates model design.

To address the above issues, we propose \textit{Switchable Whitening (SW)}.
%
SW provides a general form that integrates different whitening techniques (\eg BW, IW), as well as standardization techniques (\eg BN, IN and LN). SW controls the ratio of each technique by learning their importance weights.
%
It is able to select appropriate normalizers with respect to various vision tasks, as shown in Fig.\ref{Fig1}(b).
%
%
%
For example, semantic segmentation prefers BW and BN, while IW is mainly chosen to address image diversity in image classification.
Compared to semantic segmentation, domain adaptation selects more IW and IN, which alleviates domain discrepancy in CNN features.
In image style transfer, IW dominates to handle image style variance.

SW can be inserted into advanced CNN architectures and effectively boosts their performances.
Owing to the rich statistics and selectivity of SW, models trained with SW consistently outperform other counterparts in a number of popular benchmarks, such as CIFAR-10/100~\cite{krizhevsky2009learning} and ImageNet~\cite{deng2009imagenet} for image classification, ADE20K~\cite{zhou2017scene} and Cityscapes~\cite{cordts2016cityscapes} for semantic segmentation, domain adaptation between GTA5~\cite{Richter_2016_ECCV} and Cityscapes, and image style transfer on COCO~\cite{lin2014microsoft}.
For example, when using ResNet50~\cite{he2016deep} for ImageNet, ADE20K, and Cityscapes, as well as using VGG16~\cite{simonyan2014very} for domain adaptation, SW significantly outperforms the BN-based baselines by 1.51\%, 3.2\%, 4.1\%, and 3.0\% respectively.

SW serves as a useful tool for analyzing the characteristics of these whitening or standardization techniques.
This work answers two questions: 
(1) \textit{Is IW beneficial for high-level vision tasks like classification and domain adaptation?} 
(2) \textit{Is standardization still necessary when whitening is presented?}
Our experiments suggest that (1) IW works extremely well for handling image appearance diversity and reducing domain gap, giving rise to better performance in high-level vision tasks;
(2) Using BW+IW in SW performs comparably well compared to using all the normalizers mentioned above in SW, indicating that full whitening generally works well, and the requirement for standardization is marginal when whitening is presented.
%

%
%
%

Overall, our \textbf{contributions} are summarized as follows. 
(1) We propose Switchable Whitening (SW), which unifies existing whitening and standardization methods in a general form and learns to switch among them during training.
(2) SW adapts to various tasks and is used as a new building block in advanced CNNs. We show that SW outperforms its counterparts in multiple challenging benchmarks.
%
(3) SW could be used as a tool to analyze the effects and characteristics of different normalization methods, and the interactions between whitening and standardization.
%
We will make the code of SW available and hope it would deepen our understanding on various normalization methods.

%% file: sections/relatedwork.tex

\section{Related Work}


\noindent\textbf{Normalization.} Existing normalization techniques generally performs \textit{standardization}.
For example, Batch Normalization (BN)~\cite{ioffe2015batch} \textit{centers} and \textit{scales} activations using the mean and variance estimated over a mini-batch, accelerating training and enhancing generalization.
In contrast, Instance Normalization (IN)~\cite{ulyanov2017improved} and Layer Normalization (LN)~\cite{ba2016layer} standardize activations with statistics computed over each individual channel and all channels of a layer respectively.
IN is mainly used in image generation~\cite{huang2017arbitrary,ulyanov2017improved} while LN has been proved beneficial for training recurrent neural networks~\cite{ba2016layer}.
The above three normalizers are combined in Switchable Normalization (SN)~\cite{luo2018differentiable} that learns the ratio of each one.
The combination of BN and IN is also explored in IBN-Net~\cite{pan2018two} and Batch-Instance Normalization~\cite{nam2018batch}.
Besides, there have been other attempts to improve BN for small batch sizes such as Group Normalization~\cite{wu2018group}, Batch Renormalization~\cite{ioffe2017batch}, and Batch Kalman Normalization~\cite{wang2018batch}.
All these normalization methods perform centering and scaling to the activations, whereas the correlation between activations remains, leading to sub-optimal optimization efficiency.
%
Our work provides a general form that integrates both whitening and standardization techniques, having SN as a special case.


\noindent\textbf{Whitening.} Another paradigm towards improving optimization is \textit{whitening}.
Desjardins \etal ~\cite{desjardins2015natural} proposes Natural Neural Network, which implicitly whitens the activations to improve conditioning of the Fisher Information Matrix.
%
This improves optimization efficiency of deep neural networks.
Decorrelated Batch Normalization (DBN)~\cite{lei2018decorrelated} whitens features using covariance matrix computed over a mini-batch.
It extends BN by decorrelating features.
In this paper, we refer to DBN as Batch Whitening (BW) for consistency.
Moreover, in the field of image style transfer, whitening and coloring operations are used to manipulate the image appearance~\cite{li2017universal,siarohin2018whitening}.
This is because the appearance of an individual image is well encoded in the covariance matrix of its features.
We call whitening of an individual image as instance whitening (IW).
%
In this work, we make the first attempt to apply IW in high-level vision tasks like image classification and semantic segmentation.



%
%
%
%
%
%

%% file: sections/SW.tex
\section{Switchable Whitening (SW)}

We first present a general form of whitening as well as standardization operations, and then introduce SW.

\subsection{A General Form}

Our discussion is mainly based on CNNs, where the data have four dimensions.
Let \(\mathbf{X}\in\mathbb{R}^{C \times NHW} \) be the data matrix of a mini-batch, where \(N, C, H, W\) indicate the number of samples, number of channels, height, and width respectively.
Here \(N\), \(H\) and \(W\) are viewed as a single dimension for convenience.
Let matrix \(\mathbf{X}_n \in \mathbb{R}^{C \times HW} \) be the \(n\)th sample in the mini-batch, where \(n \in \{1, 2,..., N\}\).
Then the whitening transformation \(\phi : \mathbb{R}^{C \times HW} \to \mathbb{R}^{C \times HW} \) for a sample \(\mathbf{X}_n\) could be formulated as
\begin{align}
\phi(\mathbf{X}_n) = \bm{\Sigma}^{-1/2}(\mathbf{X}_n - \bm{\mu} \cdot \mathbf{1}^T)
\end{align}
where \(\bm{\mu}\) and \(\bm{\Sigma}\) are the mean vector and the covariance matrix calculated from the data, and \(\mathbf{1}\) is a column vector of all ones.
%
Note that different whitening methods could be achieved by calculating \(\bm{\mu}\) and \(\bm{\Sigma}\) using different sets of pixels.
We discuss them in detail as below.
%

\textbf{Batch Whitening (BW)}.
In BW~\cite{lei2018decorrelated}, the statistics are calculated in a mini-batch.
Thus 
\begin{gather}
\bm{\mu}_{bw} = \frac{1}{NHW} \mathbf{X} \cdot \mathbf{1}  \nonumber \\
\bm{\Sigma}_{bw} = \frac{1}{NHW}(\mathbf{X} - \bm{\mu} \cdot \mathbf{1}^T)(\mathbf{X} - \bm{\mu} \cdot \mathbf{1}^T)^T + \epsilon \mathbf{I}
\end{gather}
where \(\epsilon > 0\) is a small positive number to prevent a singular \(\bm{\Sigma}_{bw}\).
In this way, the whitening transformation \(\phi\) whitens the data of the entire mini-batch, \ie, \(\phi(\mathbf{X})\phi(\mathbf{X})^T = \mathbf{I}\).
%

\textbf{Instance Whitening (IW)}.
In contrast, for IW~\cite{li2017universal}, \(\bm{\mu}\) and \(\bm{\Sigma}\) are calculated within each individual sample,
\begin{gather}
\bm{\mu}_{iw} = \frac{1}{HW} \mathbf{X}_n \cdot \mathbf{1}  \nonumber \\
\bm{\Sigma}_{iw} = \frac{1}{HW}(\mathbf{X}_n - \bm{\mu} \cdot \mathbf{1}^T)(\mathbf{X}_n - \bm{\mu} \cdot \mathbf{1}^T)^T + \epsilon \mathbf{I}
\end{gather}
for \textit{n} in \{1, 2, ..., N\}.
IW whitens each samples separately, \ie, \(\phi(\mathbf{X}_n)\phi(\mathbf{X}_n)^T = \mathbf{I}\). 
~\\

Note that Eq.(1) also naturally incorporates standardization operations as its special cases.
In the covariance matrix \(\bm{\Sigma}\), the diagonal elements are the variance for each channel, while the off-diagonal elements are the correlation between channels.
Therefore, by simply setting the off-diagonal elements to zeros, the left multiplication of \(\bm{\Sigma}^{-1/2}\) equals to dividing the standard variance, so that Eq.(1) becomes standardization.

\textbf{Batch Normalization (BN)}.
BN\cite{ioffe2015batch} centers and scales data using the mean and standard deviation of a mini-batch.
Hence its mean is the same as in BW \ie, \(\bm{\mu}_{bn} = \bm{\mu}_{bw}\).
As discussed above, since BN does not decorrelate data, the covariance matrix becomes \(\bm{\Sigma}_{bn} = diag(\bm{\Sigma}_{bw})\), which is a diagonal matrix that only preserves the diagonal of \(\bm{\Sigma}_{bw}\).

\textbf{Instance Normalization (IN)}.
Similarly, in IN~\cite{ulyanov2017improved} we have \(\bm{\mu}_{in} = \bm{\mu}_{iw}\) and \(\bm{\Sigma}_{in} = diag(\bm{\Sigma}_{iw})\).

\textbf{Layer Normalization (LN)}.
LN~\cite{ba2016layer} uses the mean and variance of all channels in a sample to normalize.
Let \(\mu_{ln}\) and \(\sigma_{ln}\) denote the mean and the variance, 
then \(\bm{\mu}_{ln} = \mu_{ln} \mathbf{1} \) and \(\bm{\Sigma}_{ln} = \sigma_{ln} \mathbf{I} \).
In practice \(\mu_{ln}\) and \(\sigma_{ln}\) could be calculated efficiently from $\bm{\mu}_{in}$ and $\bm{\Sigma}_{in}$ using the results in \cite{luo2018differentiable}.
~\\


In Eq.(1), the inverse square root of the covariance matrix is typically calculated by using ZCA whitening,
\begin{align}
\bm{\Sigma}^{-1/2} = \mathbf{D} \bm{\Lambda}^{-1/2} \mathbf{D}^{T}
\end{align}
where \(\bm{\Lambda} = diag(\sigma_1, ..., \sigma_c)\) and \(\mathbf{D} = [\bm{d}_1, ..., \bm{d}_c]\) are the eigenvalues and the eigenvectors of \(\bm{\Sigma}\), \ie, \(\bm{\Sigma} = \mathbf{D} \bm{\Lambda} \mathbf{D}^{T} \), which is obtained via eigen decomposition.

So far we have formulated different whitening and normalization transforms in a general form.
In the next section, we introduce switchable whitening based on this formulation.

\subsection{Formulation of SW}

For a data sample \(\mathbf{X}_n\), a natural way to unify the aforementioned whitening and standardization transforms is to combine the mean and covariance statistics of those methods, and perform whitening using this unified statistics, giving rise to
~\\
\begin{align}
SW(\mathbf{X}_n) = \hat{\bm{\Sigma}}^{-1/2}(\mathbf{X}_n - \hat{\bm{\mu}} \cdot \mathbf{1}^T)
\end{align}
\begin{fleqn}
\begin{align}
\text{where} \qquad\ 
\hat{\bm{\mu}} = \sum_{k \in \Omega}\omega_{k} \bm{\mu}_{k}, \quad
\hat{\bm{\Sigma}} = \sum_{k \in \Omega}\omega_{k}^{\prime} \bm{\Sigma}_{k}
\end{align}
\end{fleqn}
Here \(\Omega\) is a set of statistics estimated in different ways.
In this work, we mainly focus on two cases, \ie, \(\Omega = \) \{bw, iw\} and \(\Omega = \) \{bw, iw, bn, in, ln\}, where the former switches between two whitening methods, while the later incorporates both whitening and standardization methods.
\(\omega_k\) are importance ratios to switch among different statistics.
In practice, \(\omega_k\) are generated by the corresponding control parameters \(\lambda_k\) via softmax function, \ie, \(\omega_k = \frac{e^{\lambda_k}}{\sum_{z \in \Omega} e^{\lambda_z}}\).
%
And \(\omega_{k}^{\prime}\) are defined similarly using another group of control parameters \(\lambda_{k}^{\prime}\).
This relieves the constraint of consistency between mean and covariance, which is a more general form.

%
Note that the above formulation incorporates SN~\cite{luo2018differentiable} as its special case by letting \(\Omega = \) \{bn, in, ln\}.
Our formulation is more flexible and general in that it takes into account the whole covariance matrix rather than only the diagonal.
This provides the possibility of producing decorrelated features, giving rise to either better optimization conditioning or style invariance.
SW could be easily extended to incorporate some other normalization methods like Batch Renormalization~\cite{ioffe2017batch} or Group Normalization~\cite{wu2018group}, which is out of the scope of this work.


\subsection{Training and Inference}

Switchable Whitening could be inserted extensively into a convolutional neural network (CNN).
Let $\mathrm{\Theta}$ be a set of parameters of a CNN, and $\mathrm{\Phi}$ be a set of importance weights in SW.
The importance weights are initialized uniformly, \eg $\lambda_k = 1$.
During training, $\mathrm{\Theta}$ and $\mathrm{\Phi}$ are optimized jointly by minimizing a loss function $\mathcal{L} (\mathrm{\Theta}, \mathrm{\Phi})$ using back-propagation.
%
The forward calculation of our proposed SW is presented in Algorithm \ref{alg:forward} while the backward pass is presented in Appendix.
For clearance, we use \(\Omega = \) \{bw, iw\} as an illustrative example.

In the training phase, \(\bm{\mu}_{bw}\) and \(\bm{\Sigma}_{bw}\) are calculated within each mini-batch and used to update the running mean and running covariance as in Line 7 and 8 of Algorithm \ref{alg:forward}.
During inference, the running mean and the running covariance are used as $\bm{\mu}_{bw}$ and $\bm{\Sigma}_{bw}$, while $\bm{\mu}_{iw}$ and $\bm{\Sigma}_{iw}$ are calculated independently for each sample.

In practice, the scale and shift operations are usually added right after the normalization or whitening transform to enhance the model's representation capacity.
For SW, we follow this design to introduce scale and shift parameters $\gamma$ and $\beta$ as in BN.

\begin{algorithm}[t!]
	\caption{Forward pass of SW for each iteration.}
	\label{alg:forward}
	{\fontsize{9}{9} \selectfont
		\begin{algorithmic}[1]
			\State {\bfseries Input:} mini-batch inputs \scalebox{0.9}{${\mathbf{X}\in\mathbb{R}^{C \times NHW}}$}, where the \(n\)th sample in the batch is \scalebox{0.9}{\({\mathbf{X}_n \in \mathbb{R}^{C \times HW} }\)}, \scalebox{0.9}{\( n \in \{1, 2,..., N\}\)}; importance weights \scalebox{0.9}{$\lambda_{k}$} and \scalebox{0.9}{$\lambda_{k}^{\prime}$}, \scalebox{0.9}{\( k \in \{bw,iw\} \)}; expected mean \scalebox{0.9}{$\bm{\mu}_{E}$} and expected covariance \scalebox{0.9}{$\bm{\Sigma}_{E}$}. \\
			{\bfseries Hyperparameters:} \(\epsilon\), running average momentum \(\alpha\). \\
			{\bfseries Output:} the whitened activations \{\scalebox{0.9}{\({\mathbf{\hat{X}}_n, n = 1, 2,..., N}\)}\}.
			\State calculate: \scalebox{0.8}[0.9]{\( \omega_{bw}, \omega_{iw} = Softmax( \lambda_{bw}, \lambda_{iw}), \omega_{bw}^{\prime}, \omega_{iw}^{\prime} = Softmax( \lambda_{bw}^{\prime}, \lambda_{iw}^{\prime})\)}.
			\State calculate: \scalebox{0.9}{\(\bm{\mu}_{bw} = \frac{1}{NHW} \mathbf{X} \cdot \mathbf{1}\)}.
			\State calculate: \scalebox{0.9}{\(\bm{\Sigma}_{bw} = \frac{1}{NHW}(\mathbf{X} - \bm{\mu} \cdot \mathbf{1}^T)(\mathbf{X} - \bm{\mu} \cdot \mathbf{1}^T)^T + \epsilon \mathbf{I}\)}.
			\State update: \scalebox{0.9}{\(\bm{\mu}_{E} \gets (1 - \alpha) \bm{\mu}_{E} + \alpha\bm{\mu}_{bw} \)}.
			\State update: \scalebox{0.9}{\(\bm{\Sigma}_{E} \gets (1 - \alpha) \bm{\Sigma}_{E} + \alpha\bm{\Sigma}_{bw} \)}.
			\For{$n=1$ {\bfseries to} $N$}
			\State calculate: \scalebox{0.9}{\(\bm{\mu}_{iw}^{(n)} = \frac{1}{HW} \mathbf{X}_n \cdot \mathbf{1}\)}.
			\State calculate: \scalebox{0.9}{\(\bm{\Sigma}_{iw}^{(n)} = \frac{1}{HW}(\mathbf{X}_n - \bm{\mu} \cdot \mathbf{1}^T)(\mathbf{X}_n - \bm{\mu} \cdot \mathbf{1}^T)^T + \epsilon \mathbf{I}\)}.
			\State calculate: \scalebox{0.9}{\(\bm{\hat{\mu}}_n = \sum_{k}\omega_{k} \bm{\mu}_{k}^{(n)}\), \(\bm{\hat{\Sigma}}_n = \sum_{k}\omega_{k}^{\prime} \bm{\Sigma}_{k}^{(n)}\), \(k \in \{bw,iw\}\)}.
			\State execute eigenvalue decomposition: \scalebox{0.9}{\(\bm{\hat{\Sigma}}_n = \mathbf{D} \bm{\Lambda} \mathbf{D}^{T} \)}.
			\State calculate ZCA-whitening matrix: \scalebox{0.9}{\(\mathbf{U}_n = \mathbf{D} \bm{\Lambda}^{-1/2} \mathbf{D}^{T}\)}.
			\State calculate ZCA-whitened output: \scalebox{0.88}{\(\mathbf{\hat{X}}_n = \mathbf{U}_n(\mathbf{X}_n - \hat{\bm{\mu}}_n \cdot \mathbf{1}^T)\)}.
			\EndFor
		\end{algorithmic}
	}
\end{algorithm}

\subsection{Accelerates SW via Newton's Iteration}

In practice, the GPU implementation of singular value decomposition (SVD) in current deep learning frameworks are inefficient, leading to much slower training and inference.
To address this issue, we could resort to an alternative way to calculate $\hat{\bm{\Sigma}}^{-1/2}$, which is to use Newton's iteration, as in IterNorm~\cite{huang2019iterative}.
Following~\cite{huang2019iterative}, we normalize $\hat{\bm{\Sigma}}$ via $\hat{\bm{\Sigma}}_N = \hat{\bm{\Sigma}} / tr(\hat{\bm{\Sigma}})$.
Then calculate $\hat{\bm{\Sigma}}_{N}^{-1/2}$ via the following iterations:
\begin{align}
\begin{cases}
\mathbf{P}_0 = \mathbf{I} \\
\mathbf{P}_k = \frac{1}{2} (3\mathbf{P}_{k-1} - \mathbf{P}^{3}_{k-1}\bm{\hat{\Sigma}}_N), \ k=1,2,...,T
\end{cases}
\end{align}
where $T$ is the iteration number, and $\mathbf{P}_k$ will converge to $\hat{\bm{\Sigma}}_{N}^{-1/2}$.
Finally, we have $\hat{\bm{\Sigma}}^{-1/2} = \hat{\bm{\Sigma}}_{N}^{-1/2} / \sqrt{tr(\hat{\bm{\Sigma}})}$.
In this work, we set $T = 5$, which produces similar performance with the SVD version.

\subsection{Analysis and Discussion}
We have introduced the formulation and training of SW.
Here we discuss some of its important properties and analyze its complexity.

\noindent\textbf{Instance Whitening for Appearance Invariance. }
In style transfer, researchers have found that image appearance information (\ie color, contrast, style \etc) is well encoded in the covariance matrix of features produced by CNNs~\cite{li2017universal}.
In this work, we take the first attempt to induce appearance invariance by leveraging IW, which is beneficial for domain adaptation or high-level vision tasks like classification or semantic segmentation.
Although IN also introduces invariance by standardizing each sample separately, the difference in correlation could be easily enlarged in highly non-linear deep neural networks.
In IW, features of different samples are not only standardized but also whitened individually, giving rise to the same covariance matrix, \ie, identity matrix.
Therefore, IW has better invariance property than IN.

\noindent\textbf{Switching between Whitening and Standardization.}
Our formulation of SW makes it possible to switch between whitening and standardization.
For example, considering \(\Omega = \) \{bw, bn\}, \ie, \(\hat{\bm{\Sigma}} = \omega_{bw}\bm{\Sigma}_{bw} + \omega_{bn}\bm{\Sigma}_{bn}\), \((\omega_{bw}+\omega_{bn}=1)\).
As $\omega_{bn}$ grows larger, the diagonal of $\hat{\bm{\Sigma}}$ would remain the same, while the off-diagonal would be weaken.
This would make the features less decorrelated after whitening.
This is beneficial when the extent of whitening requires careful adjustment, which is an important issue of BW as pointed out in \cite{lei2018decorrelated}.

\noindent\textbf{Group SW.}
Huang \etal~\cite{lei2018decorrelated} uses group whitening to reduce complexity and to address the inaccurate estimation of large covariance matrices.
%
%
In SW we follow the same design, \ie, the features are divided into groups along the channel dimension and SW is performed for each group.
The importance weights $\lambda_k$ could be shared or independent for each group.
In this work we let groups of a layer share the same $\lambda_k$ to simplify discussion.

\setlength{\tabcolsep}{4pt}
\begin{table}[h]
	\begin{center}
		\caption{Comparisons of computational complexity. \(N, C, H, W\) are the number of samples, number of channels, height, and width of the input tensor respectively. $G$ denotes the number of channels for each group in group whitening.}
		\label{complexity}
		\resizebox{7.5cm}{1.1cm}{
			\begin{tabular}{c|cc}
				\hline
				\multirow{2}{*}{Method} & \multicolumn{2}{c}{Computational complexity} \\ \cline{2-3} 
				& w/o group             & w/ group             \\ \hline
				BN,IN,LN,SN             & $O(NCHW)$                 &  $O(NCHW)$          \\
				BW                      & $O(C^2\text{max}(NHW,C))$  &  $O(CG\text{max}(NHW,G))$  \\
				IW                      & $O(NC^2\text{max}(HW,C))$  &  $O(NCG\text{max}(HW,G))$  \\
				SW                      & $O(NC^2\text{max}(HW,C))$  &  $O(NCG\text{max}(HW,G))$  \\ \hline
			\end{tabular}
		}
	\end{center}
\vspace{-5pt}
\end{table}
\setlength{\tabcolsep}{1.4pt}

\noindent\textbf{Complexity Analysis.}
%
%
The computational complexities for different normalization methods are compared in Table \ref{complexity}.
The flop of SW is comparable with IW.
And applying group whitening could reduce the computation by $C/G$ times.
Usually we have $HW > G$, thus the computation cost of SW and BW would be roughly the same (\ie, $O(CGNHW)$).


%% file: sections/experiments.tex
\begin{figure*}[t!]
	\centering
	\includegraphics[width=17.5cm]{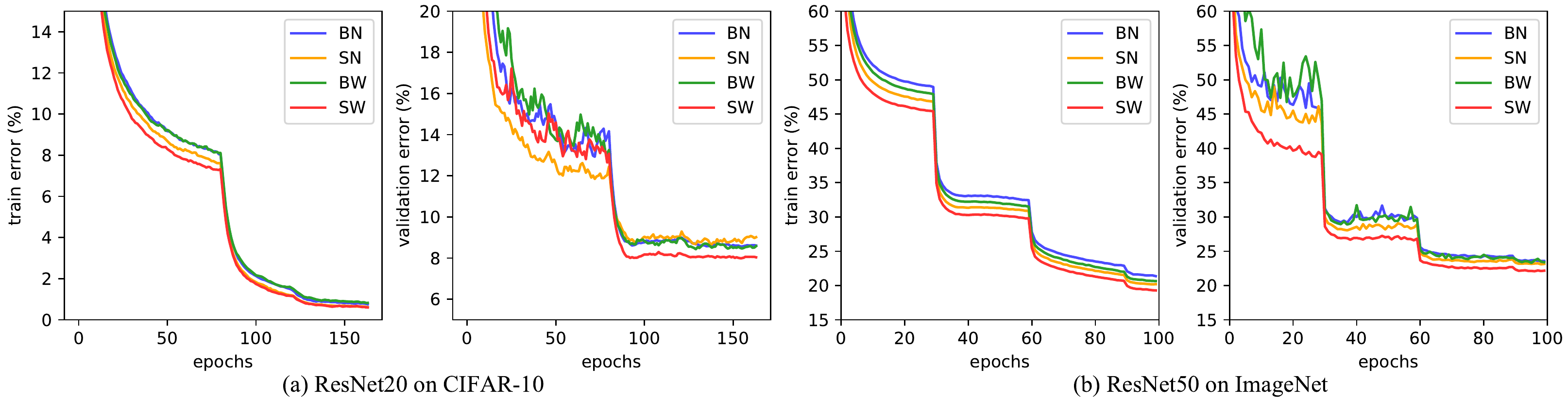}
	\vspace{-10pt}
	\caption{Training and validation error curve on CIFAR-10 and ImageNet. Models with different normalization methods are reported. Here SW has \(\Omega = \) \{bw, iw\}.}
	\label{error_curve}
	\vspace{-5pt}
\end{figure*}

\section{Experiments}

%
We evaluate SW on image classification (CIFAR-10/100, ImageNet), semantic segmentation (ADE20K, Cityscapes), domain adaptation (GTA5, Cityscapes), and image style transfer (COCO).
%
%
For each task, SW is compared with previous normalization methods.
We also provide results of instance segmentation in the Appendix.

\subsection{Classification}
%

\setlength{\tabcolsep}{4pt}
\begin{table}[!t]
	\begin{center}
		\caption{Test errors (\%) on CIFAR-10/100 and ImageNet validation sets~\cite{krizhevsky2009learning}. For each model, we evaluate different normalization or whitening methods. $\text{SW}^{a}$ and $\text{SW}^{b}$ correspond to \(\Omega = \) \{bw, iw\} and \(\Omega = \) \{bw, iw, bn, in, ln\} respectively. Results on CIFAR are averaged over 5 runs.}
		\label{cifar}
		\resizebox{8.4cm}{1.85cm}{
			\begin{tabular}{l|l|lllll}
				\Xhline{2\arrayrulewidth}
				Dataset                    & Method      & BN    &  SN    &  BW    & $\text{SW}^{a}$  & $\text{SW}^{b}$  \\ \hline\hline
				\multirow{4}{*}{CIFAR-10}  & ResNet20   & 8.45  &  8.34  &  8.28  & \textbf{7.64} & 7.75 \\ 
				& ResNet44   & 7.01  &  6.75  &  6.83  & \textbf{6.27} & 6.35 \\           
				& ResNet56   & 6.88  &  6.57  &  6.62  & \textbf{6.07} & 6.25  \\ 
				& ResNet110  & 6.21  &  5.97  &  5.99  & \textbf{5.69} & 5.78  \\ \hline
				\multirow{2}{*}{CIFAR-100} & ResNet20  & 32.09  &  32.28  & 32.44 & 31.00 & \textbf{30.87}  \\ 
				& ResNet110 & 27.32  & 27.25  &  27.76  & 26.64  &  \textbf{26.48}  \\ \hline
				\multirow{2}{*}{ImageNet} & ResNet50 (top1)	& 23.58 & 23.10 & 23.31 & 22.10 & \textbf{22.07} \\
				& ResNet50 (top5) & 7.00 & 6.55  & 6.72 & 5.96 & \textbf{5.91} \\ \Xhline{2\arrayrulewidth}
			\end{tabular}
		}
	\end{center}
	\vspace{-16pt}
\end{table}
\setlength{\tabcolsep}{1.4pt}


CIFAR-10, CIFAR-100~\cite{krizhevsky2009learning} and ImageNet~\cite{deng2009imagenet} are standard image classification benchmarks.
%
Our training policies and settings are the same as in ~\cite{he2016deep}.

\noindent\textbf{Implementation}.
We evaluate different normalization methods based on standard ResNet~\cite{he2016deep}.
Note that introducing whitening after all convolution layers of ResNet is redundant and would incur a high computational cost, as also pointed out in ~\cite{lei2018decorrelated}.
Hence we replace part of the BN layers in ResNet to the desired normalization layers.
For CIFAR, we apply SW or other counterparts after the 1\textit{st} and the \{4n\}\textit{th} (n = 1,2,3,...) convolution layers.
%
%
%
%
%
%
And for ImageNet, the normalization layers considered are those at the 1\textit{st} and the \{6n\}\textit{th} (n = 1,2,3,...) layers.
The residual blocks with 2048 channels are not considered to save computation.
%
%
More discussions for such choices could be found in supplementary material.

The normalization layers studied here are BN, SN, BW, and SW.
For SW, we consider two cases: \(\Omega = \) \{bw, iw\} and \(\Omega = \) \{bw, iw, bn, in, ln\}, which are denoted as $\text{SW}^{a}$ and $\text{SW}^{b}$ respectively.
In all experiments, we adopt group whitening with group size $G = 16$ for SW and BW.
Since \cite{luo2018normalization} shows that applying early stop to the training of SN reduces overfitting, we stop the training of SN and SW at the 80\textit{th} epoch for CIFAR and the 30\textit{th} epoch for ImageNet.

\noindent\textbf{Results}.
The results are given in Table.~\ref{cifar} and the training curves are shown in Fig.~\ref{error_curve}.
In both datasets, $\text{SW}^{a}$ and $\text{SW}^{b}$ show better results and faster convergence than BN, SN, and BW over various network depth.
Specifically, with only 7 $\text{SW}^{b}$ layers, the top1 and top5 error of ResNet50 on ImageNet is significantly reduced by 1.51\% and 1.09\%.
This performance is comparable with the original ResNet152 which has 5.94\% top5 error.
%
%

Our results reveal that combining different normalization methods in a suitable manner surpasses every single normalizer.
For example, the superiority of $\text{SW}^{b}$ over SN attributes to the better optimization conditioning brought out by whitening.
And the better performance of $\text{SW}^{a}$ over BW shows that instance whitening is beneficial as it introduces style invariance.
Moreover, $\text{SW}^{a}$ and $\text{SW}^{b}$ perform comparably well, which indicates that full whitening generally performs well, and the need for standardization is marginal while whitening is presented.

\noindent\textbf{Discussions}.
SW has two groups of importance weights \(\lambda_{k}\) and \(\lambda_{k}^{\prime}\).
We observe that allowing \(\lambda_{k}\) and \(\lambda_{k}^{\prime}\) to share weight produces slightly worse results.
For example, ResNet20 has 8.17\% test error when using SW with shared importance weights.
We conjecture that mean and covariance have different impacts in training, and recommend to maintain independent importance weights for mean and covariance. 

Note that IW is not reported here because it generally produces worse results due to diminished feature discrimination.
For example, ResNet20 with IW gives 12.57\% test error on CIFAR-10, which is worse than other normalization methods.
This also implies that SW borrows the benefits of different normalizers so that it could outperform any individual of them. 

\subsection{Semantic Segmentation}

\setlength{\tabcolsep}{4pt}
\begin{table}[!t]
	\begin{center}
		\caption{Results on Cityscapes and ADE20K datasets. `ss' and `ms' indicate single-scale and multi-scale test respectively.}
		\label{segmentation}
		\resizebox{7.0cm}{1.5cm}{
			\begin{tabular}{l|cc|cc}
				\Xhline{2\arrayrulewidth}
				\multirow{2}{*}{Method} & \multicolumn{2}{c|}{ADE20K} & \multicolumn{2}{c}{Cityscapes} \\ \cline{2-5} 
				& mIoU\textsubscript{ss} & mIoU\textsubscript{ms} & mIoU\textsubscript{ss} & mIoU\textsubscript{ms} \\ \hline\hline
				ResNet50-BN             & 36.6           &    37.9        & 72.1         &    73.4      \\
				ResNet50-SN             & 37.8           &    38.8        & 75.0         &    76.2      \\
				ResNet50-BW             & 35.9           &    37.8       & 72.5         &     73.7     \\
				ResNet50-$\text{SW}^{a}$ & \textbf{39.8}  &   \textbf{40.8}        & \textbf{76.2}         &    \textbf{77.1}     \\
				ResNet50-$\text{SW}^{b}$ & \textbf{39.8}  &    40.7        & 76.0     &  77.0        \\ \Xhline{2\arrayrulewidth}
			\end{tabular}
		}
	\end{center}
	\vspace{-12pt}
\end{table}
\setlength{\tabcolsep}{1.4pt}

\setlength{\tabcolsep}{4pt}
\begin{table}[!t]
	\begin{center}
		\caption{Comparison with advanced methods on the ADE20K validation set. * indicates our implementation. }
		\label{ade20k2}
		\resizebox{6.0cm}{2.2cm}{
			\begin{tabular}{l|cc}
				\Xhline{2\arrayrulewidth}
				Method       & mIoU(\%) & Pixel Acc.(\%) \\ \hline\hline
				DilatedNet~\cite{yu2015multi}  & 32.31  & 73.55  \\
				CascadeNet~\cite{zhou2016semantic}  & 34.90  & 74.52  \\
				RefineNet~\cite{lin2017refinenet} & 40.70    & -              \\
				PSPNet101~\cite{zhao2017pyramid} & 43.29    & 81.39          \\
				SDDPN~\cite{liang2018dynamic} & 43.68    & 81.13			\\
				WiderNet~\cite{wu2019wider}   & 43.73    & 81.17          \\
				PSANet101~\cite{zhao2018psanet}  & 43.77    & 81.51          \\
				EncNet~\cite{zhang2018context}  & 44.65		& 81.69			\\ \hline
				PSPNet101*                & 43.59           & 81.41         \\
				PSPNet101-$\text{SW}^{a}$ & \textbf{45.33} & \textbf{82.05} \\ \Xhline{2\arrayrulewidth}
			\end{tabular}
		}
	\end{center}
	\vspace{-15pt}
\end{table}
\setlength{\tabcolsep}{1.4pt}

We further verify the scalability of our method on ADE20K~\cite{zhou2017scene} and Cityscapes~\cite{cordts2016cityscapes}, which are standard and challenging semantic segmentation benchmarks.
%
%
%
%
We evaluate SW based on ResNet and PSPNet~\cite{zhao2017pyramid}.

\noindent\textbf{Implementation.} 
We adopt the same ResNet architecture, training setting, and data augmentation scheme as in \cite{zhao2017pyramid}.
The normalization layers considered are the 1\textit{st} and the \{3n\}\textit{th} (n = 1,2,3,...) layers except those with 2048 channels, resulting in 14 normalization layers for ResNet50.
Since overfitting is not observed in these two benchmarks, early stop is not used here.
The BN and BW involved are synchronized across multiple GPUs.

\noindent\textbf{Results.} 
Table.\ref{segmentation} reports mIoU on the validation sets of the two benchmarks.
For ResNet50, simply replacing part of BN with SW would significantly boost mIoU\textsubscript{ss} by 3.2\% and 4.1\% for ADE20K and Cityscapes respectively.
SW also notably outperforms SN and BW, which is consistent with the results of classification.

Furthermore, we show that SW could improve even the most advanced models for semantic segmentation.
We apply SW to PSPNet101~\cite{zhao2017pyramid}, and compare with other methods on the ADE20K dataset.
The results are shown in Table.\ref{ade20k2}.
Simply using some SW layers could improve the strong PSPNet by 1.74\% on mIoU.
And our final score, 45.33\%, outperforms other more advanced semantic segmentation methods like PSANet~\cite{zhao2018psanet} and EncNet~\cite{zhang2018context}.

\noindent\textbf{Computational cost.} 
While the above implementation of SW is based on SVD, they can be accelerated via Newton's iteration, as discussed in section 3.4.
As shown in Table.\ref{time}, the GPU running time is significantly reduced when using iterative whitening, while the performance is comparable to the SVD version.
Note that in this Table, the ResNet-50-$\text{SW}^{a}$ in Cityscapes has the same configuration as in ImageNet, \ie, has 7 SW layers.
Compared with the 14 layer version, this further saves computation cost, while still achieves satisfactory results.

\begin{table}[t!]
	\begin{center}
		\caption{Performance and running time of ResNet50 with different normalization layers on ImageNet and Cityscapes datasets. We report the GPU running time per iteration during training. The GPU we use is NVIDIA Tesla V100.}
		\label{time}
		\resizebox{7.8cm}{1.20cm}{
			\begin{tabular}{l|c|cc|cc}
				\Xhline{2\arrayrulewidth}
				\multicolumn{1}{c|}{\multirow{2}{*}{Method}} & \multirow{2}{*}{Whitening} & \multicolumn{2}{c|}{ImageNet} & \multicolumn{2}{c}{Cityscapes} \\ \cline{3-6} 
				&                            & error(\%)       & time(s)       & mIoU(\%)        & time(s)        \\ \hline
				ResNet50-BN            & -                          & 23.58        & 0.27           & 72.1         & 0.52            \\
				ResNet50-BW            & svd                        & 23.31        & 0.79           & 72.4         & 1.09            \\
				ResNet50-$\text{SW}^{a}$  & svd                        & 22.10        & 1.04           & 75.7         & 1.24            \\
				ResNet50-$\text{SW}^{a}$  & iterative                  & 22.07        & 0.36           & 76.0         & 0.67            \\ \Xhline{2\arrayrulewidth}
			\end{tabular}
		}
	\end{center}
	\vspace{-10pt}
\end{table}

\setlength{\tabcolsep}{4pt}
\begin{table*}[t!]
	\begin{center}
		\caption{Results of adapting GTA5 to Cityscapes. mIoU of models with different normalization layers are reported.}
		\vspace{-6pt}
		\label{DA}
		\resizebox{17.5cm}{1.35cm}{
			\begin{tabular}{l|ccccccccccccccccccc|c}
				\Xhline{2\arrayrulewidth}
				Method & \R{road} & \R{sidewalk} & \R{building} & \R{wall} & \R{fence} & \R{pole} & \R{light} & \R{sign} & \R{veg} & \R{terrain} & \R{sky} & \R{person} & \R{rider} & \R{car} & \R{truck} & \R{bus} & \R{train} & \R{mbike} & \R{bike} & mIoU \\ \hline\hline
				AdaptSetNet-BN & 88.3 & 42.7 & 74.9 & 22.0 & 14.0 & 16.5 & 17.8 & 4.2 & 83.5 & 34.3 & 72.1 & 44.8 & 1.7 & 76.9 & 18.0 & \textbf{6.7} & 0.0 & 3.0 & 0.1 & 32.7 \\
				AdaptSetNet-SN & 87.0 & 41.6 & 77.5 & 21.2 & \textbf{20.0} & \textbf{18.3} & 20.9 & \textbf{8.3} & 82.4 & 35.4 & 72.6 & 48.4 & 1.4 & 81.1 & \textbf{18.7} & 5.2 & 0.0 & 8.4 & 0.0 & 34.1 \\
				AdaptSetNet-$\text{SW}^{a}$ & \textbf{91.8} & 50.2 & 78.1 & \textbf{25.3} & 17.5 & 17.5 & \textbf{21.4} & 6.2 & 83.4 & 36.6 & 74.0 & \textbf{50.7} & \textbf{7.4} & 83.4 & 16.7 & 6.3 & 0.0 & \textbf{10.4} & \textbf{0.8} & \textbf{35.7} \\
				AdaptSetNet-$\text{SW}^{b}$ & \textbf{91.8} & \textbf{50.5} & \textbf{78.4} & 23.5 & 16.5 & 17.2 & 19.8 & 5.5 & \textbf{83.6} & \textbf{38.4} & \textbf{74.6} & 48.9 & 5.3 & \textbf{83.6} & 17.6 & 3.9 & \textbf{0.1} & 7.7 & 0.7 & 35.1 \\ \Xhline{2\arrayrulewidth}
			\end{tabular}
		}
	\end{center}
	\vspace{-15pt}
\end{table*}
\setlength{\tabcolsep}{1.4pt}

\subsection{Domain Adaptation}

The adaptive style invariance of SW making it suitable for handling appearance discrepancy between two image domains.
To verify this, we evaluate SW on domain adaptation task.
The datasets employed are the widely used GTA5~\cite{Richter_2016_ECCV} and Cityscapes~\cite{cordts2016cityscapes} datasets.
GTA5 is a street view dataset generated semi-automatically from the computer game Grand Theft Auto V (GTA5), while Cityscapes contains traffic scene images collected from the real world.
%


\noindent\textbf{Implementation.} 
We conduct our experiments based on the AdaptSegNet~\cite{tsai2018learning} framework, which is a recent state-of-the-art domain adaptation approach.
It adopts adversarial learning to shorten the discrepancy between two domains with a discriminator.
The segmentation network is DeepLab-v2~\cite{chen2018deeplab} model with VGG16~\cite{simonyan2014very} backbone.
The training setting is the same as in \cite{tsai2018learning}.

Note that the VGG16 model has five convolutional groups, where the number of convolution layers for these groups are \{2,2,3,3,3\}.
We add SW or its counterparts after the first convolution layer of each group, and report the results using different normalization layers.

\begin{figure}[t!]
	\centering
	\includegraphics[width=8.0cm]{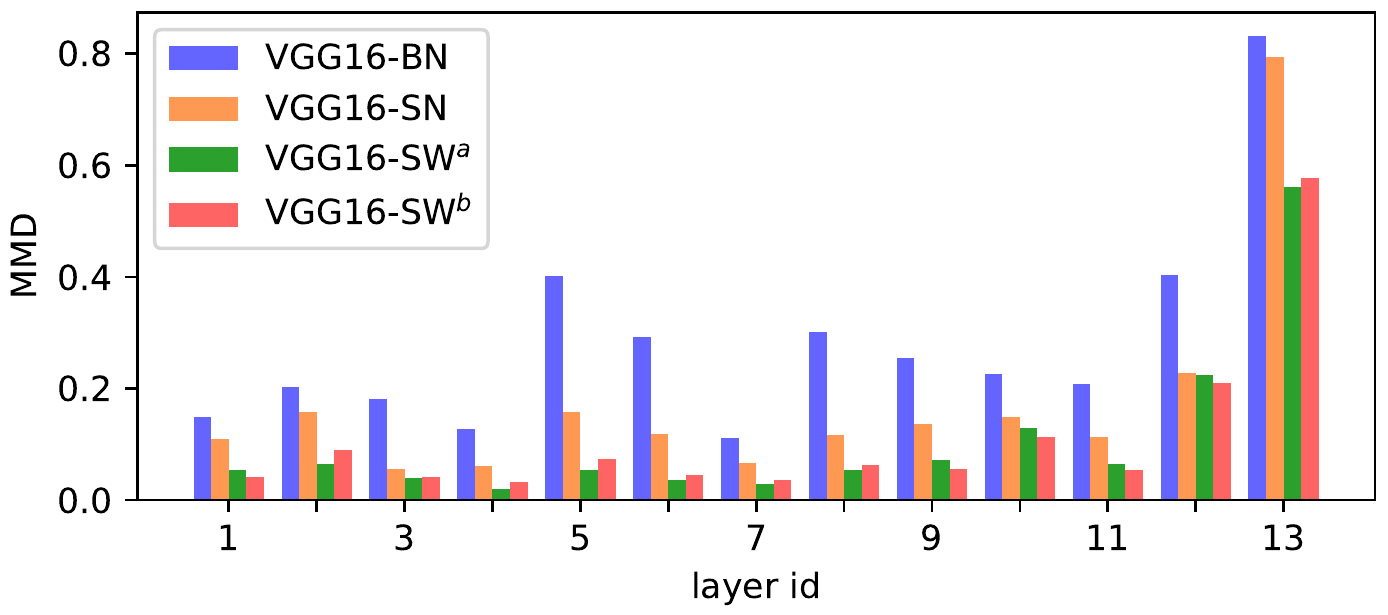}
	\caption{MMD distance between Cityscapes and GTA5.}
	\label{MMD}
	\vspace{-10pt}
\end{figure}

\begin{figure*}[t!]
	\centering
	\includegraphics[width=15.5cm, height=3.5cm]{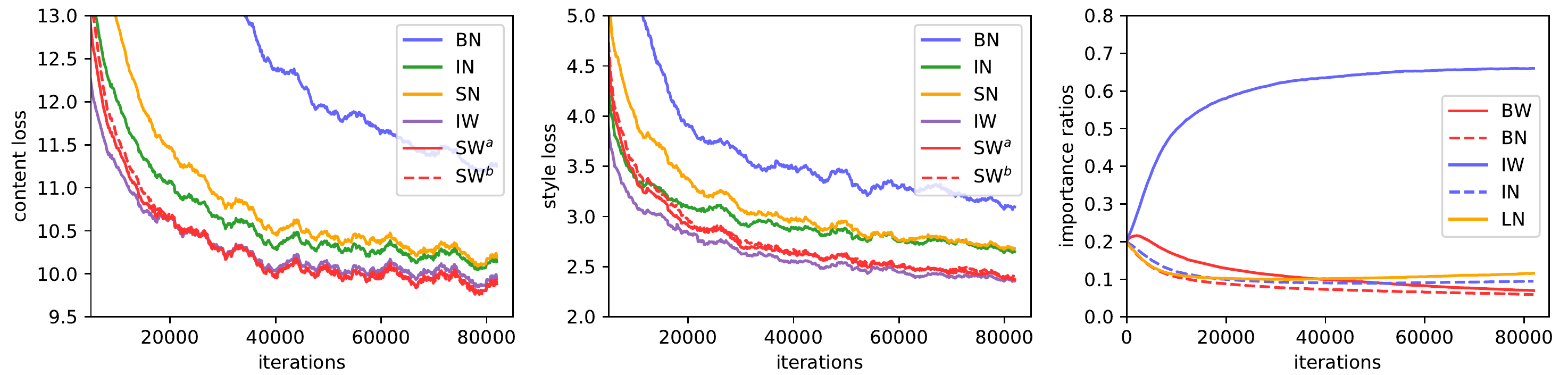}
	\caption{Training loss in style transfer and the learned importance ratios of $\text{SW}^{b}$. The importance ratios are averaged over all $\text{SW}^{b}$ layers in the image stylizing network. }
	\label{style_loss}
	\vspace{-12pt}
\end{figure*}

\begin{figure}[t!]
	\centering
	\includegraphics[width=8.4cm]{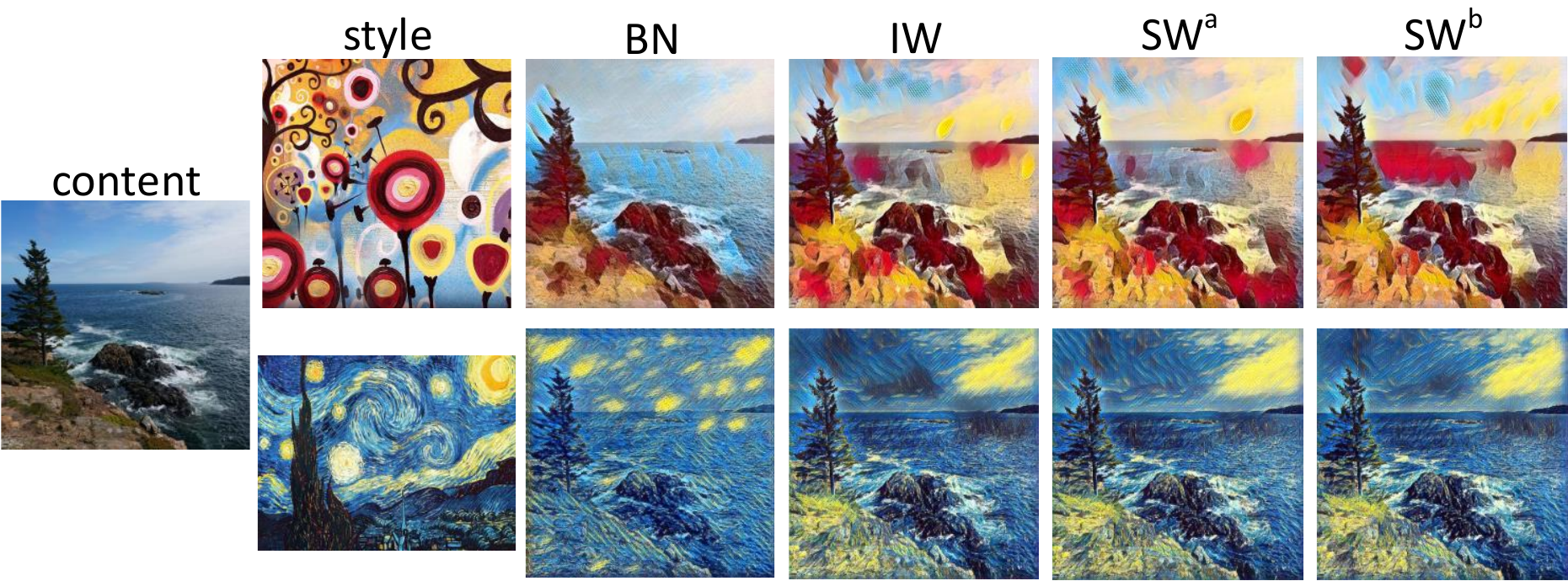}
	\vspace{-5pt}
	\caption{Visualization of style transfer using different normalization layers.}
	\label{visualize}
	\vspace{-10pt}
\end{figure}

\noindent\textbf{Results.} 
Table.\ref{DA} reports the results of adapting GTA5 to Cityscapes.
The models with SW achieve higher performance when evaluated on a different image domain.
Particularly, compared with BN, and SN, $\text{SW}^{a}$ improves the mIoU by 3.0\%, and 1.6\% respectively.

To understand how SW performs better under cross-domain evaluation, we analyze the maximum mean discrepancy (MMD)\cite{gretton2012kernel} of deep features between the two datasets.
MMD is a commonly used metric for evaluating domain discrepancy.
Specifically, we use the MMD with Gaussian kernels as in \cite{li2017mmd}.
We calculate the MMD for features of the first 13 layers in VGG16 with different normalization layers.
The results are shown in Fig.\ref{MMD}.
Compared with BN and SN, SW significantly reduces MMD for both shallow and deep features.
This shows that the IW introduced effectively reduces domain discrepancy in the CNN features, making the model easier to generalize to other data domains.

\subsection{Image Style Transfer}

Thanks to the rich statistics, SW could work not only in high-level vision tasks, but also in low-level vision tasks like image style transfer.
To show this, we employ a popular style transfer algorithm~\cite{johnson2016perceptual}.
It has an image stylizing network trained with the content loss and style loss calculated by a loss network.
The MS-COCO dataset~\cite{lin2014microsoft} is used as content images while the style images selected are \textit{candy} and \textit{starry night}.
We follow the same training policy as in \cite{johnson2016perceptual}, and adopt different normalization layers for the image stylizing network.

\noindent\textbf{Results.}
The training loss curve is shown in Fig.\ref{style_loss}.
As revealed in former works, IW and IN perform better than BN.
Besides, we observe that IW has smaller content loss and style loss than IN, which verifies that IW works better in manipulating image style.
Although SW converges slower than IW at the beginning, it soon catches up with IW as SW learns to select IW as the normalizer.
Moreover, SW has smaller content loss than IW when the training converges, as BW preserves important content information.

Qualitative examples of style transfer using different normalization layers are shown in Fig.\ref{visualize}.
BN produces poor stylization images, while IW gives satisfactory results.
SW works comparably well with IW, showing that SW is able to select appropriate normalizer according to the task.
More examples are provided in supplementary material.

\subsection{Analysis on SW}

\begin{figure*}[t!]
	\centering
	\includegraphics[width=13.5cm, height=3.4cm]{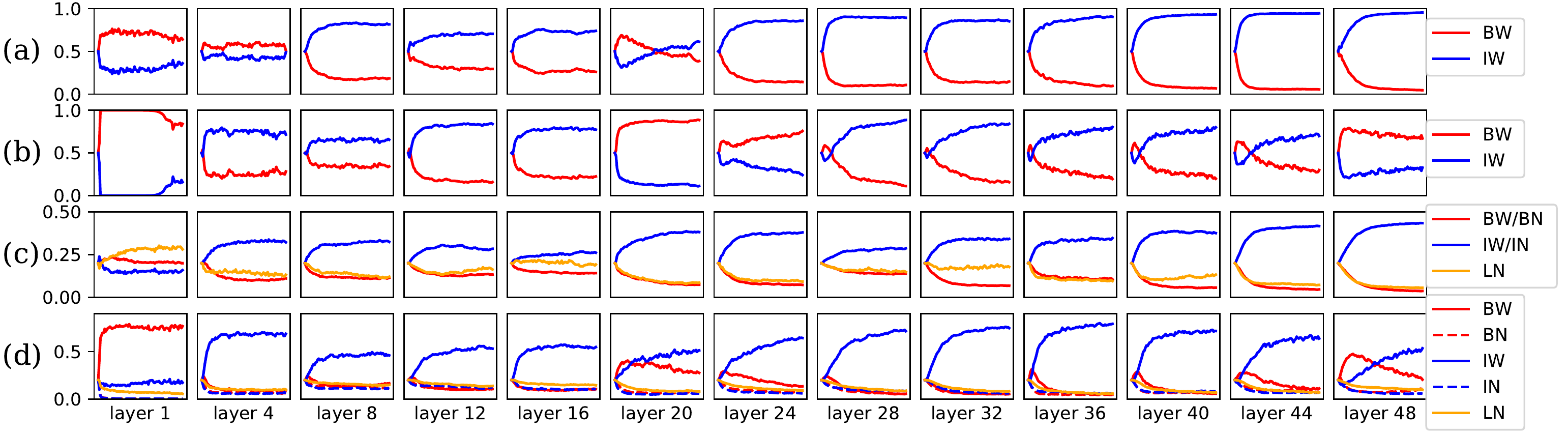}
	\caption{Learning curve of importance weights in ResNet56 on CIFAR-10. (a) and (b) show \(\omega_k\) and \(\omega_{k}^{\prime}\) in SW with \(\Omega = \) \{bw, iw\}. (c) and (d) correspond to \(\omega_k\) and \(\omega_{k}^{\prime}\) in SW with \(\Omega = \) \{bw, iw, bn, in, ln\}.}
	\label{CIFAR10_weight}
\end{figure*}

\begin{figure*}[t!]
	\centering
	\includegraphics[width=14.0cm, height=3.4cm]{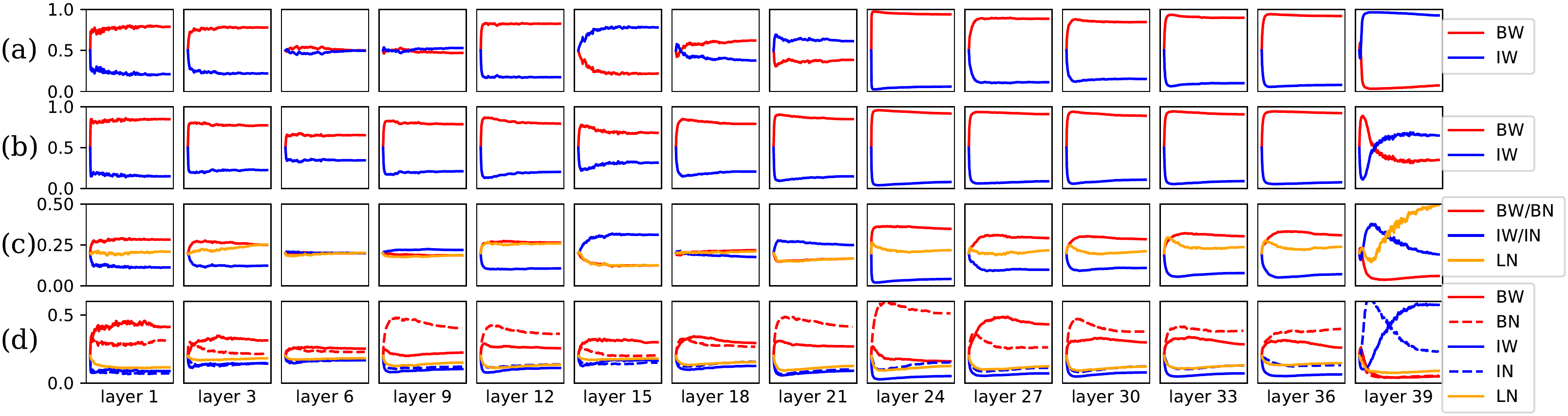}
	\caption{Learning curve of importance weights in ResNet50 on Cityscapes. (a)(b)(c)(d) have the same meanings as in Fig.\ref{CIFAR10_weight}.}
	\label{cs_weight}
\end{figure*}

\begin{figure*}[t!]
	\centering
	\includegraphics[width=11.2cm]{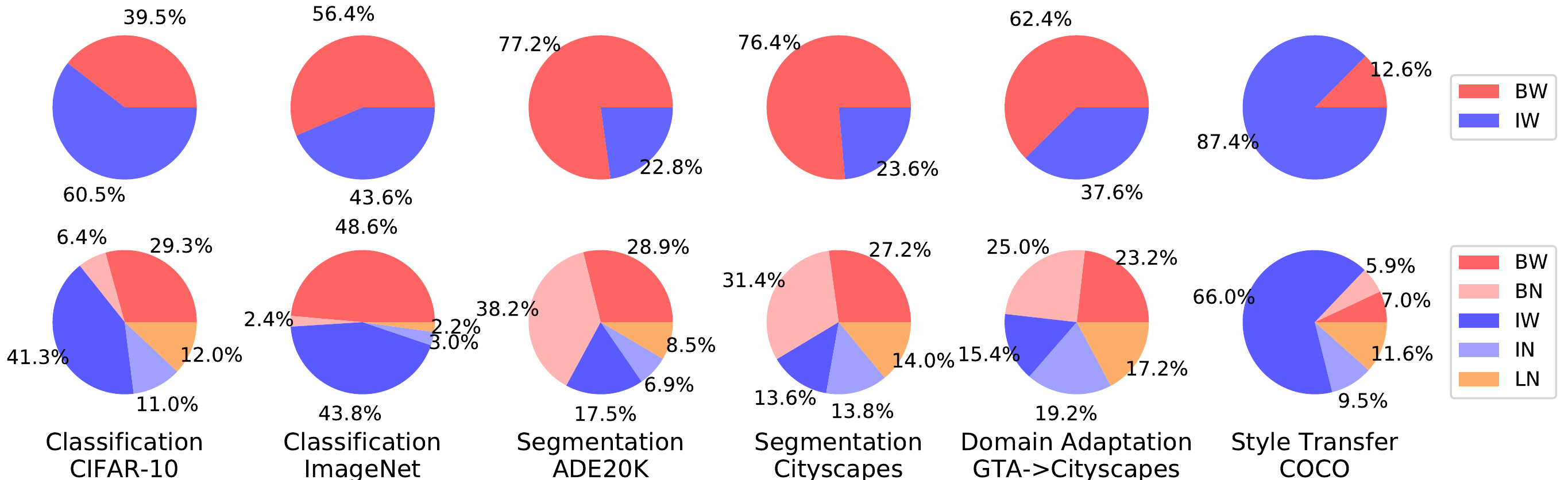}
	\caption{Learned importance ratios of SW in various tasks. Above and below correspond to \(\Omega = \) \{bw, iw\} and \(\Omega = \) \{bw, iw, bn, in, ln\} respectively.}
	\label{ratio}
	\vspace{-8pt}
\end{figure*}

In order to understand the behavior of SW, in this section we study its learning dynamics and the learned importance ratios.

\noindent\textbf{Learning Dynamics.}
The importance ratios of SW is initialized to have uniform values, i.e. 0.5 for \(\Omega = \) \{bw, iw\} and 0.2 for \(\Omega = \) \{bw, iw, bn, in, ln\}.
To see how the ratios of SW in different layers change during training, we plot the learning curves of \(\omega_k\) and \(\omega_{k}^{\prime}\) in Fig.\ref{CIFAR10_weight} and Fig.\ref{cs_weight}. 
It can be seen that the importance ratios shift quickly at the beginning and gradually become stable.
There are several interesting observations.
(1) The learning dynamics vary across different tasks.
In CIFAR-10, SW mostly selects IW and occasionally selects BW, while in Cityscapes BW or BN is mostly chosen.
(2) The learning behaviours of SW across different layers tend to be distinct rather than homogeneous.
For example, in Fig.\ref{cs_weight} (a), SW selects IW for layer \{15, 21, 39\}, and BW for the rest except for layer \{6, 9\} where the ratios keep uniform.
%
(3) The behaviors of \(\omega_k\) and \(\omega_{k}^{\prime}\) are mostly coherent and sometimes divergent.
For instance, in layer \{15, 21\} of Fig.\ref{cs_weight}, \(\omega_k\) chooses IW while \(\omega_{k}^{\prime}\) chooses BW or BN.
This implies that \(\bm{\mu}\) and \(\bm{\Sigma}\) are not necessarily have to be consistent, as they might have different impacts in training.
%

\noindent\textbf{Importance Ratios for Various Tasks.}

We further analyze the learned importance ratios in SW for various tasks, as shown in Fig.\ref{ratio}.
The results are obtained by taking average over the importance ratios \(\omega_{k}^{\prime}\) of all SW layers in a CNN.
The models are ResNet50 for ImageNet, ADE20K, and Cityscapes, ResNet44 for CIFAR-10, VGG16 for GTA5$\rightarrow$Cityscapes, and a ResNet-alike network for style transfer as in \cite{johnson2016perceptual}.
Both \(\Omega = \) \{bw, iw\} and \(\Omega = \) \{bw, iw, bn, in, ln\} are reported.

We make the following remarks: 
(1) For semantic segmentation, SW chooses mainly BW and BN, and partially the rest, while in classification more IW are selected.
This is because the diversity between images is higher in classification datasets than in segmentation datasets.
Thus more IW is required to alleviate the intra-dataset variance. 
(2) Semantic segmentation on Cityscapes tends to produce more IW and IN under domain adaptation setting than in the normal setting.
Since domain adaptation introduces a domain discrepancy loss, more IW and IN would be beneficial for reducing the feature discrepancy between the two domains, \ie, GTA5 and Cityscapes.
(3) In image style transfer, SW switches to IW aggressively.
This phenomenon is consistent with the common knowledge that IW is well suited for style transfer, as image level appearance information is well encoded in the covariance of CNN features.
Our experiments also verify that IW is a better choice than IN in this task.

%% file: sections/conclusion.tex
\section{Conclusion}

In this paper, we propose Switchable Whitening, which integrates various whitening and standardization techniques in a general form.
SW adapts to various tasks by learning to select appropriate normalizers in different layers of a CNN.
Our experiments show that SW achieves consistent improvements over previous normalization methods in a number of computer vision tasks, including classification, segmentation, domain adaptation, and image style transfer.
Investigation of SW reveals the importance of leveraging different whitening methods in CNNs.
%
We hope that our findings in this work would benefit other research fields and tasks that employ deep learning.

%% file: sections/supplementary.tex
\clearpage

\setcounter{section}{0}
\renewcommand\thesection{\Alph{section}}

\section*{\fontsize{15}{15}\selectfont Appendix}

This appendix provides (1) back-propagation of SW, (2) discussion for our network configurations, (3) results for instance segmentation, and (4) some style transfer visualization results.

\section{Back-propagation of SW}

The backward pass of our proposed SW is presented in Algorithm \ref{alg:backward}.
\vspace{-8pt}
\begin{algorithm}[h]
	\caption{Backward pass of SW for each iteration.}
	\label{alg:backward}
	{\fontsize{9}{9} \selectfont
		\begin{algorithmic}[1]
			\State {\bfseries Input:} mini-batch gradients respect to whitened outputs \{\scalebox{0.9}{\(\frac{\partial L}{\partial \mathbf{\hat{X}}_n}, n = 1, 2,..., N\)}\}. Other auxiliary data from respective forward pass.
			\State {\bfseries Output:} the gradients respect to the inputs \{\scalebox{0.9}{\(\frac{\partial L}{\partial \mathbf{X}_n}, n = 1, 2,..., N\)}\};
			the gradients respect to the importance weights \scalebox{0.9}{$\frac{\partial L}{\partial \lambda_k}$} and \scalebox{0.9}{$\frac{\partial L}{\partial \lambda_k^{\prime}}$}, \scalebox{0.9}{\( k \in \{bw,iw\} \)}.
			\For{$n=1$ {\bfseries to} $N$}
			\State calculate \scalebox{0.9}{\(\frac{\partial L}{\partial \bm{\hat{\Sigma}}_n}\)} using results below.
			\State calculate \scalebox{0.9}{\(\frac{\partial L}{\partial \bm{\hat{\mu}}_n}\)} using results below.
			\EndFor
			\For{$n=1$ {\bfseries to} $N$}
			\State calculate \scalebox{0.9}{
				\( \frac{\partial L}{\partial \mathbf{X}_n} = \frac{\partial L}{\partial \mathbf{\hat{X}}_n} \mathbf{U}_n + ( \frac{\omega_{bw}}{NHW} \sum_{i=1}^{N} \frac{\partial L}{\partial \bm{\hat{\mu}_i}} + \frac{\omega_{iw}}{HW} \frac{\partial L}{\partial \bm{\hat{\mu}_n}} )
				\)
			}
			\scalebox{0.9}{
				\( + \lbrack \frac{2\omega_{bw}^{\prime}(\mathbf{X_n} - \bm{\mu}_{bw})^T}{NHW} \sum_{i=1}^{N} (\frac{\partial L}{\partial \bm{\hat{\Sigma}_i}})_{sym} + \frac{2\omega_{iw}^{\prime}(\mathbf{X}_n - \bm{\mu}_{iw})^T}{HW} (\frac{\partial L}{\partial \bm{\hat{\Sigma}_n}})_{sym} \rbrack 
				\)
			}
			\EndFor
			\State calculate: \linebreak
			\scalebox{0.85}[0.9]{
				\( \frac{\partial L}{\partial \lambda_{bw}} = \omega_{bw}(1 - \omega_{bw}) \sum_{n=1}^{N} (\frac{\partial L}{\partial \bm{\hat{\mu}}_n} \bm{\mu}_{bw} ) - \omega_{iw}\omega_{bw} \sum_{n=1}^{N} (\frac{\partial L}{\partial \bm{\hat{\mu}}_n} \bm{\mu}_{iw}^{(n)} ) \)
			}
			\scalebox{0.85}[0.9]{
				\( \frac{\partial L}{\partial \lambda_{iw}} = \omega_{iw}(1 - \omega_{iw}) \sum_{n=1}^{N} (\frac{\partial L}{\partial \bm{\hat{\mu}}_n} \bm{\mu}_{iw}^{(n)} ) - \omega_{bw}\omega_{iw} \sum_{n=1}^{N} (\frac{\partial L}{\partial \bm{\hat{\mu}}_n} \bm{\mu}_{bw} ) \)
			}
			\scalebox{0.82}[0.9]{
				\( \frac{\partial L}{\partial \lambda_{bw}^{\prime}} = \omega_{bw}^{\prime}(1 - \omega_{bw}^{\prime}) \sum_{n=1}^{N} \langle \frac{\partial L}{\partial \bm{\hat{\Sigma}}_n}, \bm{\Sigma}_{bw} \rangle_F - \omega_{iw}^{\prime}\omega_{bw}^{\prime} \sum_{n=1}^{N} \langle \frac{\partial L}{\partial \bm{\hat{\Sigma}}_n}, \bm{\Sigma}_{iw}^{(n)} \rangle_F \)
			}\footnotemark \linebreak
			\scalebox{0.82}[0.9]{
				\( \frac{\partial L}{\partial \lambda_{iw}^{\prime}} = \omega_{iw}^{\prime}(1 - \omega_{iw}^{\prime}) \sum_{n=1}^{N} \langle \frac{\partial L}{\partial \bm{\hat{\Sigma}}_n}, \bm{\Sigma}_{iw}^{(n)} \rangle_F - \omega_{bw}^{\prime}\omega_{iw}^{\prime} \sum_{n=1}^{N} \langle \frac{\partial L}{\partial \bm{\hat{\Sigma}}_n}, \bm{\Sigma}_{bw} \rangle_F \)
			}
		\end{algorithmic}
	}
\end{algorithm}
\vspace{-8pt}

\footnotetext{$\langle \rangle_F$ denotes Frobenius inner product.}

In the Next, we derive \(\frac{\partial L}{\partial \bm{\hat{\Sigma}}_n}\) and \(\frac{\partial L}{\partial \bm{\hat{\mu}}_n}\) in the line 4, 5 of Algorithm \ref{alg:backward}, where we start with the forward pass of ZCA whitening.

\noindent\textbf{Forward Pass.}
Let \(\mathbf{X}\in\mathbb{R}^{C \times HW} \) be a sample of a mini-batch.
Here the subscript $n$ is omitted for clearance.
Given the integrated mean $\hat{\bm{\mu}}$ and the integrated covariance $\hat{\bm{\Sigma}}$ in SW, the ZCA whitening is as follows:
\begin{align}
\bm{\hat{\Sigma}} &= \mathbf{D} \bm{\Lambda} \mathbf{D}^{T}  \\
\mathbf{V} &= \bm{\Lambda}^{-1/2} \mathbf{D}^{T}	\\
\mathbf{\tilde{X}} &= \mathbf{V} (\mathbf{X} - \hat{\bm{\mu}} \cdot \mathbf{1}^T) \\
\mathbf{\hat{X}} &= \mathbf{D} \mathbf{\tilde{X}}
\end{align}
where $\mathbf{V}$ and $\mathbf{\tilde{X}}$ are intermediate variables for clarity in derivation.

\noindent\textbf{Back-propagation.}
Based on the chain rule and the results in \cite{ionescu2015training,lei2018decorrelated}, $\frac{\partial L}{\partial \bm{\hat{\Sigma}}}$ and $\frac{\partial L}{\partial \bm{\hat{\mu}}}$ can be calculated as follows:
\begin{align}
\frac{\partial L}{\partial \mathbf{\tilde{X}}} &= \frac{\partial L}{\partial \mathbf{\hat{X}}} \mathbf{D}	\\
\frac{\partial L}{\partial \mathbf{V}} &= \frac{\partial L}{\partial \mathbf{\tilde{X}}}^{T} (\mathbf{X} - \hat{\bm{\mu}} \cdot \mathbf{1}^T)^T	\\
\frac{\partial L}{\partial \bm{\Lambda}} &= \frac{\partial L}{\partial \mathbf{V}} \mathbf{D} (-\frac{1}{2} \bm{\Lambda}^{-3/2})	\\
\frac{\partial L}{\partial \mathbf{D}} &= \frac{\partial L}{\partial \mathbf{V}} \bm{\Lambda}^{-1/2}) + \frac{\partial L}{\partial \mathbf{\hat{X}}} \mathbf{\tilde{X}}^T	\\
\frac{\partial L}{\partial \bm{\hat{\Sigma}}} &= \mathbf{D} \{[\mathbf{K}^T \odot (\mathbf{D}^T \frac{\partial L}{\partial \mathbf{D}})] + (\frac{\partial L}{\partial \bm{\Lambda}})_{diag} \} \mathbf{D}^T		\\
\frac{\partial L}{\partial \bm{\hat{\mu}}} &= \frac{\partial L}{\partial \mathbf{\tilde{X}}} (-\mathbf{V})
\end{align}
where $L$ is the loss calculated via a loss function, $\mathbf{K} \in \mathbb{R}^{C \times C}$ is a 0-diagonal matrix with $\mathbf{K}_{ij} = \frac{1}{\sigma_i - \sigma_j} [i \neq j]$, and $\odot$ is element-wise matrix multiplication.
%

\section{Discussion for Network Configurations}

In our experiments, we replace part of the BN layers in ResNet to SW layers to save computation and to reduce redundancy.
In this section we discuss the network configurations in detail.

\noindent\textbf{CIFAR-10/100.}
For CIFAR, the ResNet has two convolution layers in a residual module. 
Thus we apply SW or other counterparts after the 1\textit{st} and the \{4n\}\textit{th} (n = 1,2,3,...) convolution layers.
For example, in ResNet20, the normalization layers considered are the \{1,4,8,12,16\}\textit{th} layers.
We consider the 1\textit{st} layer because \cite{lei2018decorrelated} shows that it is effective to conduct whitening there.
The last layer is not considered because it is a classifier where normalization is not needed.

\noindent\textbf{ADE20K and Cityscapes.}
For semantic segmentation, the ResNet50 has a bottleneck architecture with a period of three layers, where the second layer provides a compact embedding of the input features.
Therefore we apply SW after the second convolution layer of the bottleneck.
Then the normalization layers considered are those at the 1\textit{st} and the \{3n\}\textit{th} (n = 1,2,3,...) layers.
The residual blocks with 2048 channels are not considered to save computation, which also follows the rule in \cite{pan2018two} that instance normalization should not be added in deep layers.
Thus in ResNet50, the normalization layers considered are the \{1,3,6,...,39\}\textit{th} layers, containing 14 layers in total.

\noindent\textbf{ImageNet.}
And for ImageNet, the network configuration is similar to ResNet50 in semantic segmentation, except that we consider the 1\textit{st} and the \{6n\}\textit{th} (n = 1,2,3,...) layers to further save computation.
Thus the normalization layers considered are the \{1,6,12,...,36\}\textit{th} layers, containing 7 layers in total.

\begin{figure*}[t!]
	\centering
	\includegraphics[width=16.0cm]{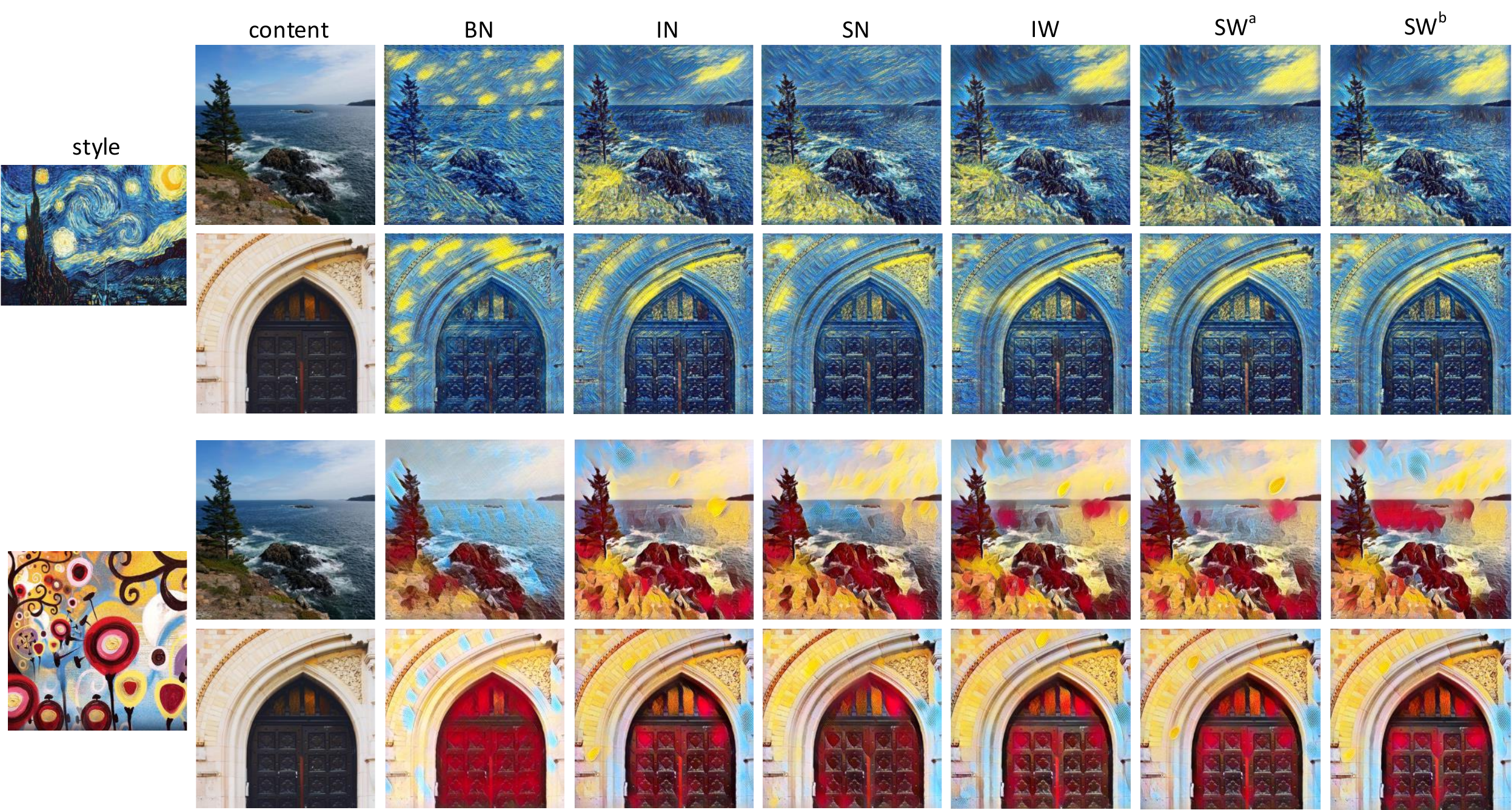}
	\caption{Visualization of style transfer using different normalization layers.}
	\label{style}
	\vspace{-6pt}
\end{figure*}

\section{Instance Segmentation}

\begin{table}[h]
	\begin{center}
		\caption{Mask R-CNN using ResNet50 and FPN with $2\times$ LR schedule.}
		\label{instanceseg}
		\resizebox{5.5cm}{1.30cm}{
\begin{tabular}{cc|cc}
	\Xhline{2\arrayrulewidth}
	Backbone & ~ FPN \& Head & $\text{AP}_{box}$  & $\text{AP}_{mask}$  \\ \hline
	FrozenBN & -         & 38.5 & 35.1 \\
	SyncBN   & SyncBN    & 39.6 & 35.6 \\
	GN       & GN        & 39.6 & 35.8 \\
	SN       & SN        & 41.0 & 36.5 \\
	$\text{SW}^{a}$ & SyncBN & \textbf{41.2} & \textbf{37.0} \\ \Xhline{2\arrayrulewidth}
\end{tabular}
		}
	\end{center}
	\vspace{-13pt}
\end{table}

We further provide results on instance segmentation, where Mask-RCNN~\cite{he2017mask} and COCO dataset~\cite{lin2014microsoft} are used to evaluate our method, and the implementation is based on mmdetection~\cite{chen2019mmdetection}.
We replace 7 normalization layers of the ResNet50 backbone with SW following the same way as in ImageNet, while the rest normalization layers of the backbone, FPN, and detection/mask head are SyncBN.
The SW layers are synchronized across multiple GPUs.
As shown in Fig.\ref{instanceseg}, SW significantly outperforms SyncBN and GN~\cite{wu2018group}, and also outperforms SN reported by~\cite{luo2018differentiable}, which replaces all normalization layers to SN.

\section{Style Transfer Results}

Fig.\ref{style} provides visualization examples for image style transfer, where results of stylizing network with different normalization techniques are shown.
It can be observed that the results of BN are worse than those of other methods, and SW produces comparably well stylizing images with IW.
This shows that SW well adapts to the image style transfer task.